\documentclass[lettersize,journal]{IEEEtran}
\usepackage{amsmath,amssymb,amsfonts}
\usepackage{algorithm}
\usepackage{array}
\usepackage[caption=false,font=normalsize,labelfont=sf,textfont=sf]{subfig}
\usepackage{textcomp}
\usepackage{stfloats}
\usepackage{url}
\usepackage{verbatim}
\usepackage{graphicx}
\usepackage{cite}
\usepackage{wrapfig}
\usepackage{xcolor}
\usepackage{cleveref}

\usepackage{algpseudocode}
\usepackage{amsmath}

\crefname{equation}{}{}

\begin{document}

\title{Synthetic Data Generation for Residential Load Patterns via Recurrent GAN and Ensemble Method}

\author{Xinyu~Liang,
        Ziheng~Wang,
        and~Hao~Wang,~\IEEEmembership{Member,~IEEE}
\thanks{This work was supported in part by the Australian Research Council (ARC) Discovery Early Career Researcher Award (DECRA) under Grant DE230100046. (Corresponding author: Hao Wang.)}
\thanks{X. Liang, Z. Wang, and H. Wang are with Department of Data Science and AI, Faculty of Information Technology and Monash Energy Institute, Monash University, Melbourne, VIC 3800, Australia.} 
}

\markboth{IEEE Transactions on Instrumentation and Measurement}%
{Liang \MakeLowercase{\textit{et al.}}: Synthetic Data Generation for Residential Load Patterns}


\maketitle

\begin{abstract}
Generating synthetic residential load data that can accurately represent actual electricity consumption patterns is crucial for effective power system planning and operation. The necessity for synthetic data is underscored by the inherent challenges associated with using real-world load data, such as privacy considerations and logistical complexities in large-scale data collection. In this work, we tackle the above-mentioned challenges by developing the Ensemble Recurrent Generative Adversarial Network (ERGAN) framework to generate high-fidelity synthetic residential load data. ERGAN leverages an ensemble of recurrent Generative Adversarial Networks, augmented by a loss function that concurrently takes into account adversarial loss and differences between statistical properties. Our developed ERGAN can capture diverse load patterns across various households, thereby enhancing the realism and diversity of the synthetic data generated. Comprehensive evaluations demonstrate that our method consistently outperforms established benchmarks in the synthetic generation of residential load data across various performance metrics including diversity, similarity, and statistical measures. The findings confirm the potential of ERGAN as an effective tool for energy applications requiring synthetic yet realistic load data. We also make the generated synthetic residential load patterns publicly available\footnote{https://github.com/AdamLiang42/ERGAN-Dataset}.
\end{abstract}

\begin{IEEEkeywords}
Residential electricity load, synthetic load generation, generative adversarial network.
\end{IEEEkeywords}

\section{Introduction}

\IEEEPARstart{T}{he} crucial role of individual user load data in power system operation, such as distribution system operation and home energy management, underscores the necessity of having access to such data for effective decision-making. Specifically, load data, which characterizes users' electricity demand, can be utilized for various studies, including residential load forecasting~\cite{zhao2023spatial}, load disaggregation~\cite{feng2021nonintrusive,ghosh2021improved,wang2022fed}, electricity theft and anomaly detection~\cite{qi2022novel,chu2022anomaly}, and electric vehicle and changing detection~\cite{jaramillo2020supervised,martin2023non,ke2024divide}. These studies are crucial for understanding energy consumption behaviors and facilitating subsequent applications such as renewable energy integration, home energy management, demand response, transactive energy, distribution grid planning, and voltage control~\cite{anvari2022data}. The operational effectiveness of these applications relies largely on accessing and utilizing such load data~\cite{darwazeh2022review,wang2024ai}.

Nevertheless, obtaining accurate and granular load data of residential users presents unique challenges. Unlike the industrial and commercial sectors with their relatively predictable load profiles, residential load is influenced by a myriad of factors - weather conditions, geographical location, household size, and diverse appliance usage patterns \cite{anvari2022data,nti2020electricity}. With the increasing adoption of residential renewable energy systems and smart home technologies, these patterns have become even more complex, underscoring the need for high-quality residential load data. However, accessing and collecting real-world residential load data are often hindered by stringent privacy concerns and formidable logistical complexities \cite{fernandez2022privacy}. As a result, the development of methods for generating synthetic residential load data, which accurately mimics real-world patterns without infringing upon privacy norms, is rapidly gaining attention.

In response to the growing need for residential load data, numerous studies have sought to develop load data generation methods. A prominent approach within this body of research is employs physical modeling methods \cite{marszal2016household,subbiah2013activity,ding2016analytical,klemenjak2020synthetic,el2020data,diao2017modeling}. These models typically account for various physical factors that influence residential energy consumption, such as the type and number of electric appliances, the insulation characteristics of the housing, and the behavior patterns of the residents. More specifically, these models often simulate the operation of each individual appliance within a household based on user behaviors, weather conditions, appliance use, and other housing characteristics. The resulting consumption patterns of each individual component are then aggregated to form the overall load profile of the household. Although this method can yield highly accurate and detailed synthetic load data, it is inherently complex in collecting information concerning driving factors and requires substantial computational resources. For example, collecting detailed appliance and user behavior data can be challenging and often infringe on privacy, presenting a significant limitation to this approach.

To tackle these limitations, traditional statistical or probabilistic methods have emerged as a more feasible alternative to physical modeling, particularly for directly simulating aggregate-level load data \cite{duque2021conditional,zufferey2018generating,labeeuw2013residential,wagner2016modeling,uhrig2014statistical}. By leveraging the statistical characteristics of historical load data, this approach generates synthetic profiles that are representative of residential electricity consumption. In essence, these methods transcribe aggregated load data into synthetic equivalents, bypassing the needs for detailed appliance, housing, or occupant specific information. This approach not only preserves privacy but also simplifies the data collection process. Various techniques, such as Markov chain models, Copulas models, and Gaussian mixture models, fall into this category and have exhibited satisfactory performance in extracting key characteristics and patterns of load data. However, traditional statistical or probabilistic methods have their limitations. Owing to the multifaceted factors influencing electricity consumption, a key shortcoming of this approach lies in its inability to accurately capture complex and non-linear relationships inherent in residential load data. The oversimplification caused by these methods often leads to synthetic data, which is statistically similar but lacks fidelity to real-world patterns. Therefore, while these traditional methods address some of the issues encountered in physical modeling, they fall short of comprehensively capturing the complexities of residential load profiles. 

Recognizing the limitations of traditional statistical and probabilistic methods in directly modeling residential-level load, there has been a shift towards deep learning based techniques in recent years. Such techniques have demonstrated successful applications in a broad spectrum of energy data related topics, such as renewable energy scenario generation \cite{chen2018model}, renewable power forecasting \cite{huang2022time}, battery lifecycle forecasting\cite{ardeshiri2022gated}, wind farm turbine fault detection
\cite{liu2021sparse}, residential load forecasting \cite{zhang2019scenario}, and system level load data generation \cite{wang2020generating,asre2022synthetic}. The distinguishing feature of these methodologies is their innate ability to learn and model complex data patterns. However, the direct application of these methods for residential load data generation presents unique challenges. The distinct granularity, dynamics, and heterogeneity are inherent in residential load profiles necessitate tailored solutions that can effectively capture this intricacy and faithfully generate synthetic data without compromising on diversity or realism. A recent study has successfully employed the Auxiliary Classifier Generative Adversarial Network (ACGAN) to generate residential load data \cite{gu2019gan}. Despite the relative advancement offered by ACGAN, it has still been identified to potentially compromise the diversity in the generated data. This limitation primarily stems from the dual functionality of the discriminator within the ACGAN model which is tasked with classifying real or synthetic data while simultaneously determining the data class. This conditional Generative Adversarial Network (GAN) based method can inadvertently result in limiting the generator's ability to learn from diverse data patterns that belong to the same class, thus affecting its diversity.

In response to the research gap identified above, we develop the Ensemble Recurrent Generative Adversarial Network (ERGAN) model. ERGAN effectively integrates ensemble learning and recurrent GAN architectures, aimed at effectively capturing the complexity of diverse household load patterns. In addition, ERGAN incorporates statistical properties in the loss function, alongside adversarial losses. This dual focus ensures a closer resemblance of the generated load patterns to the original distribution. This innovation enhances the realism and diversity of synthetic load data, thereby presenting a promising advancement for power system operation research relying on load data. The contributions of this paper are summarized as follows.
\begin{itemize}
    \item This work presents an effective framework, named ERGAN, to generate synthetic residential load patterns. It effectively encapsulates the complexity and diversity of residential load patterns, ensuring that the synthetic data maintains high fidelity with real-world scenarios.
    \item By leveraging the strengths of an ensemble of recurrent GANs, ERGAN diversifies and elevates the quality of the generated synthetic data. This distinctive architecture sets the ERGAN framework apart from existing studies, especially in terms of data diversity.
    \item The ERGAN framework introduces a unique loss function implementation that integrates statistical property differences along with the adversarial loss. This further ensures the generated data's alignment with the original distribution. This approach contributes significantly to the model's superior performance.
    \item We evaluate the ERGAN framework against several state-of-the-art benchmarks in residential load pattern data generation across different performance metrics, such as diversity, similarity, and statistical measures, ensuring a comprehensive assessment.
\end{itemize}

\section{Methodology}

\begin{figure*}[t]
\centerline{\includegraphics[width=1.0\textwidth]{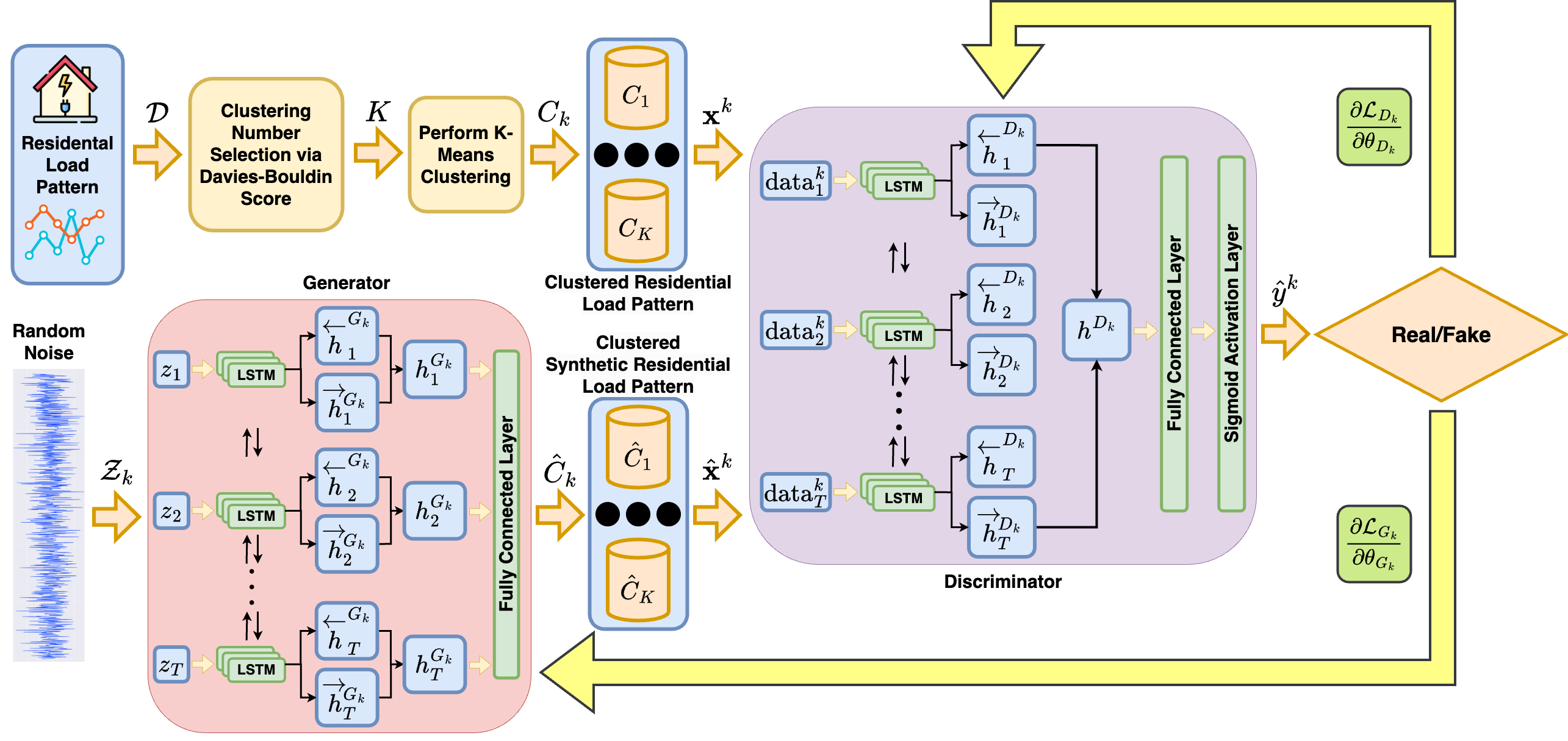}}
\caption{The ERGAN framework for generating synthetic residential load patterns via an ensemble of recurrent GANs and K-means clustering. Note that this framework involves clustering original load patterns using K-means and Davies-Bouldin score, training separate Bi-LSTM GAN models for each cluster, generating synthetic patterns from the trained generators, and combining the synthetic clustered datasets to form the final synthetic load pattern dataset while preserving cluster structures.}
\label{fig:ERGAN}
\end{figure*}

The methodology of this study is established on the synergistic integration of K-means clustering and GANs to generate synthetic residential load patterns, as shown in Figure \ref{fig:ERGAN} and Algorithm \ref{ergan_algorithm}, which we will explain in detail later. In the first phase, the data is divided into $K$ discrete clusters utilizing K-means clustering. 
For each of these $K$ clusters, a separate GAN model is trained independently to learn the data distribution specific to that cluster. This design allows the ERGAN framework to capture the unique characteristics and variabilities of different clusters more effectively than training a single GAN model on the entire dataset.
To do so, this partitioned data serves as the input for subsequent GAN models, which encompass a generator and a discriminator. Both these components employ Bi-Directional Long Short-Term Memory (Bi-LSTM) \cite{huang2015bidirectional} networks for data generation. The integrated loss function in our models combines adversarial losses with statistical property differences, thereby maintaining alignment with the original data distribution. Upon completion of the GAN models' training, the generated outputs from all models are consolidated to yield the final synthetic residential load data. The following subsections provide a more detailed exposition of each component of the ERGAN framework.
\begin{algorithm}[t]
\caption{ERGAN}\label{ergan_algorithm}
\begin{algorithmic}[1]
\State \textbf{Input:} Dataset of residential load patterns $D$
\State \textbf{Output:} Synthetic dataset $\hat{D}$

\State \textbf{Determine Optimal Number of Clusters $K$:}
\State Initialize range for $K$ (e.g., 2 to 12)
\For{each $K$ in range}
    \State Apply K-means clustering to $D$ to form $K$ clusters
    \State Calculate Davies-Bouldin Index for this $K$
\EndFor
\State Select $K$ with the lowest Davies-Bouldin Index

\State \textbf{Cluster Data using Optimal $K$:}
\State Apply K-means clustering to $D$ to create $K$ clusters $\{C_1, C_2, \ldots, C_K\}$

\State \textbf{Initialize Generative Models:}
\For{each cluster $C_k$}
    \State Generator $G_k$ with weight $\theta_{G_k}$ initialized randomly
    \State Discriminator $D_k$ with weight $\theta_{D_k}$ initialized randomly
\EndFor

\State \textbf{Train Generative Models:}
\For{each cluster $C_k$}
    \While{not converged}
        \For{each batch of data in $C_k$}
            \State $z \gets$ sample noise vector
            \State $\hat{\textbf{x}}^k \gets G_k(z)$
            \State $\textbf{x} \gets$ sample from $C_k$
            \State Update $\theta_{D_k}$ of $D_k$
            \State Update $\theta_{G_k}$ of $G_k$
        \EndFor
    \EndWhile
\EndFor

\State \textbf{Generate Synthetic Data:}
\State Initialize $\hat{D}$ as empty
\For{each $G_k$}
    \State Generate synthetic patterns $\hat{C}_k$
    \State $\hat{D} \gets \hat{D} \cup \hat{C}_k$
\EndFor

\State \textbf{Return} $\hat{D}$
\end{algorithmic}
\end{algorithm}

\subsection{Problem Formulation}
The residential load pattern generation can be formulated as a time-series generation problem. Let us consider a dataset $\mathcal{D}$ of residential load patterns, each denoted as a time-series sequence $\mathbf{x} = (x_t)_{t=1}^T$, where $T$ represents the time duration and $x_t$ represents power consumption in $t$-th time slot.

Our goal is to create a generative framework ${Gen}$ that can generate synthetic residential load patterns $\hat{\mathbf{x}} = (\hat{x}_t)_{t=1}^T$ to construct synthetic dataset $\hat{\mathcal{D}}$. More specifically, at the dataset level, the objective is to minimize the divergence between the distribution of the real load patterns $p_{\mathcal{D}}(\textbf{x})$ and the distribution of the generated load patterns $p_{\hat{\mathcal{D}}}(\hat{\textbf{x}})$, formulated as:
\begin{align}
\min_{Gen}~Div(p_{\mathcal{D}}(\textbf{x}) || p_{\hat{\mathcal{D}}}(\hat{\textbf{x}})),
\end{align}
where ${Div}$ is a divergence measure.

Since the load patterns exhibit temporal dependencies, the generative framework should capture these dependencies to generate realistic patterns. Thus, the generative model should minimize the divergence between a synthetic power consumption pattern $\hat{x}_{m}$ and original power consumption pattern $x_{m}$ at any time slot $m \in \{2,...T-1\}$ conditioned on both the previous and future time slots:
\begin{align}
    \min_{Gen}~ Div(&p_\mathcal{D}\left(x_{m}|(x_t)_{t=1}^{m-1}, (x_t)_{t=m+1}^T\right)|| \nonumber\\
    &p_{\hat{\mathcal{D}}}\left(\hat{x}_{t}|(x_t)_{t=1}^{m-1}, (x_t)_{t=m+1}^T\right)).
\end{align}

The generative framework thus can generate synthetic load patterns that not only follow the historical consumption pattern up to the current time slot but also anticipate the future consumption pattern.

\subsection{K-means Clustering and Davies-Bouldin Score for Optimal Cluster Selection}

As shown in Figure \ref{fig:ERGAN}, the construction of our ERGAN framework begins with the application of K-means clustering to the residential load pattern dataset $\mathbf{\mathcal{D}} = (\mathbf{x}_i)_{i = 1}^{N}$, where $N$ represents the total number of load patterns and $\mathbf{x}_i$ represents the $i$-th load pattern in the dataset. Our method creates an initial grouping of the load patterns into $K$ distinct clusters. By dividing the load patterns in this manner, we aim to capture the inherent structures and variances within the dataset, laying the foundation for the production of synthetic load patterns that are both diverse and representative.

For each cluster denoted as $C_k$, where $k=1,...,K$, we denote the centroid as $\mathbf{c}_k$.
The objective of the K-means clustering algorithm is to minimize the within-cluster sum of squares (WCSS), mathematically defined as:
\begin{align}
\min \sum_{k=1}^{K} \sum_{\mathbf{x}_i \in C_k} ||\mathbf{x}_i - \mathbf{c}_k||^2.
\end{align}

The clustering algorithm iteratively executes two operations: cluster assignment and centroid updating. 
In the cluster assignment step, each load pattern profile $\mathbf{x}_i$ is assigned to the cluster with the closest centroid $\mathbf{c}_j$. Mathematically, this assignment is expressed as:
    \begin{equation}
    \text{Cluster assignment:} \quad \mathbf{x}_i \in C_k, \quad k = \arg \min_{j} \|\mathbf{x}_i - \mathbf{c}_j\|^2_2,
    \end{equation}
    where $j \in \{1...K\}$ is an index used to iterate over all clusters, and $\mathbf{c}_j$ is the centroid of cluster $j$.
    
    In the centroid updating step, the position of each centroid $\mathbf{c}_k$ is recalculated based on the current members of its cluster. This step is defined as:
\begin{align}
\text{Centroid update:} &\quad \mathbf{c}_k := \frac{1}{|C_k|}\sum_{i \in C_{k}} \mathbf{x}_{i},
\end{align}
where $|C_k|$ symbolizes the number of load patterns associated with the cluster $C_k$. These two stages are iterated until a steady state is reached (the cluster assignments no longer fluctuate), or until a pre-determined maximum number of iterations has been achieved.

The selection of the number of clusters $K$ is a critical aspect of the algorithm's execution. An appropriate $K$ captures the granularity of the load patterns, therefore, to guide this selection, we employ the Davies-Bouldin (DB) index-a measurement that quantifies the average similarity between each pair of clusters \cite{davies1979cluster}. The DB index is formally defined as:
\begin{align}
DB = \frac{1}{K} \sum_{k=1}^{K} \max_{k' \neq k} \left( R_{kk'} \right),
\end{align}
where $R_{kk'}$ is a measure of the similarity between two clusters $C_k$ and $C_{k'}$, given by the formula:
\begin{align}
R_{kk'} = \frac{s_k + s_{k'}}{d_{kk'}},
\end{align}
in which $s_k$ denotes the average distance of all patterns in cluster $C_k$ to its centroid $\mathbf{c}_k$, defined as:
\begin{align}
s_k = \frac{1}{|C_k|} \sum_{\mathbf{x}_i \in C_k} ||\mathbf{x}_i - \mathbf{c}_k||,
\end{align}
and $d_{kk'}$ is the Euclidean distance between centroids $\mathbf{c}_k$ and $\mathbf{c}_{k'}$ of clusters $C_k$ and $C_{k'}$, calculated as

\begin{align}
d_{kk'} = ||\mathbf{c}_k - \mathbf{c}_{k'}||_2.
\end{align}

\subsection{Generative Model Design and Construction}

After identifying the clusters and their corresponding centroids, for each cluster $C_k$, we construct a recurrent GAN model, which consists of a generator and a discriminator.

Residential load profiles exhibit strong temporal correlations influenced by factors such as household activities, weather conditions, and energy usage habits. This results in power consumption that often depends on past consumption and potential future consumption. Bi-LSTM networks are particularly suitable for this context as they can process sequences in both forward and backward directions, enabling them to capture complex temporal dependencies present in residential load patterns. Specifically, Bi-LSTM offers a superior capacity to model dependencies between load consumption both background and forward across various time slots, making them highly proficient in generating residential load profiles. Thus, both the generator and the discriminator are implemented using Bi-LSTM networks. 

For the GAN model of the $k$-th cluster, we denote the generator as $G_k$ and the discriminator as $D_k$. For each GAN model, the generator $G_k$ takes a random noise vector $z$ sampled from a predefined noise distribution $p_z(z)$ and produces synthetic load patterns $\hat{\mathbf{x}}^{k} = (\hat{x}_{1}^{k}, \hat{x}_{2}^{k}, \dots, \hat{x}_{T}^{k})$. 

The following Equations \cref{eq:gen1,eq:gen2,eq:gen3,eq:gen4} detail the step-by-step process within $G_k$ to generate synthetic load patterns $\hat{\textbf{x}^k}$.
At each time step $t$, the generator produces a forward hidden state $\overrightarrow{\mathbf{h}}_{t}^{G_k}$ and a backward hidden state $\overleftarrow{\mathbf{h}}_{t}^{G_k}$ using the Bi-LSTM cells:
\begin{align} \label{eq:gen1}
    \overrightarrow{\mathbf{h}}_{t}^{G_k} &= \overrightarrow{\text{LSTM}}(z_t, \overrightarrow{\mathbf{h}}_{t-1}^{G_k}; \theta_{G_k}),
\end{align}
\begin{align} \label{eq:gen2}
    \overleftarrow{\mathbf{h}}_{t}^{G_k} &= \overleftarrow{\text{LSTM}}(z_t, \overleftarrow{\mathbf{h}}_{t+1}^{G_k}; \theta_{G_k}).
\end{align}

The forward hidden state $\overrightarrow{\mathbf{h}}_{t}^{G_k}$ captures the dependencies on the previous time steps, while the backward hidden state $\overleftarrow{\mathbf{h}}_{t}^{G_k}$ captures the dependencies on the future time steps.
These forward and backward hidden states are then concatenated to form the combined hidden state $\mathbf{h}_{t}^{G_k}$ at time step $t$:
\begin{equation} \label{eq:gen3}
\mathbf{h}_{t} ^{G_k}= [\overrightarrow{\mathbf{h}}_{t}^{G_k}, \overleftarrow{\mathbf{h}}_{t}^{G_k}].
\end{equation}

The combined hidden state $\mathbf{h}_{t}^{G_k}$ encapsulates both past and future dependencies, enabling the generator to generate realistic load patterns that follow the temporal dependencies present in the original data.
Finally, the load value $\hat{x}_{t}^{k}$ at time step $t$ is generated by passing the combined hidden state $\mathbf{h}_t$ through a non-linear function $f$, which is a fully connected layer:
\begin{equation} \label{eq:gen4}
\hat{x}_{t}^{k} = f(\mathbf{h}_{t}^{G_k}; \theta_{G_k}),
\end{equation}
where $\theta_{G_k}$ represents the trainable parameters of the generator network. The generator $G_k$  trained on data from cluster $C_k$, utilizes the above process to generate synthetic load patterns that are inherently associated with the corresponding cluster. The collection of synthetic load patterns is used to form the synthetic clustered dataset denoted as $\hat{C}_k$, which will be formally defined in Equation \cref{eq:cluster_gen}.

The discriminator \(D_k\), on the other hand, is designed to differentiate between real and generated load patterns. It plays a crucial adversarial role in the GAN framework, that learns to distinguish between original residential load patterns from the training data and synthetic patterns produced by the generator. The discriminator's output, presented as a probability score indicating the likelihood of the input being real, guides the generator to produce increasingly realistic synthetic data. The dynamic interaction between the generator and discriminator drives the adversarial training process, ultimately leading to the generation of high-quality synthetic load patterns. In our settings, for each time step \(t\), the input load pattern can either be real, denoted as \(x_t^k\), or synthetic, denoted as \(\hat{x}_t^k\). We define the data term \(\text{data}_t^k\) for the discriminator at each time step as follows:
\[
    \text{data}_t^k = 
    \begin{cases} 
    x_t^k & \text{for real data}, \\
    \hat{x}_t^k & \text{for synthetic data}.
    \end{cases}
\]
Using this definition, the discriminator processes \(\text{data}_t^k\) and produces both a forward hidden state \(\overrightarrow{\mathbf{h}}_t^{D_k}\) and a backward hidden state \(\overleftarrow{\mathbf{h}}_t^{D_k}\) using Bi-LSTM cells:
\begin{align}
\overrightarrow{\mathbf{h}}_{t}^{D_k} &= \overrightarrow{\text{LSTM}}(\text{data}_t^k, \overrightarrow{\mathbf{h}}_{t-1}^{D_k}; \theta_{D_k}), \\
\overleftarrow{\mathbf{h}}_{t}^{D_k} &= \overleftarrow{\text{LSTM}}(\text{data}_t^k, \overleftarrow{\mathbf{h}}_{t+1}^{D_k}; \theta_{D_k}).
\end{align}

For the discriminator $D_k$, instead of using the combined hidden states at every time step, we leverage the final forward hidden state $\overrightarrow{\mathbf{h}}_T^{D_k}$ and the final backward hidden state $\overleftarrow{\mathbf{h}}_1^{D_k}$, and concatenate them to form the combined final hidden state $\mathbf{h}^{D_k}$:
\begin{equation}
\mathbf{h}^{D_k} = [\overrightarrow{\mathbf{h}}_T^{D_k}, \overleftarrow{\mathbf{h}}_1^{D_k}].
\end{equation}

This combined hidden state $\mathbf{h}^{D_k}$ encapsulates the dependencies from both directions, capturing the temporal patterns in the entire input sequence.

Finally, the discriminator passes $\mathbf{h}^{D_k}$ through a fully connected layer with a sigmoid activation function denote as $g$ to produce the output $\hat{y}^k$:

\begin{equation}
\hat{y}^k = g(\mathbf{h}^{D_k}; \theta_{D_k}).
\end{equation}

The detailed model architecture is provided in Table \ref{tab:model_architecture} in Section~\ref{sec:result}.

\subsection{Generative Model Training}

The training process for each GAN model involves a two-player min-max game between $G_k$ and $D_k$, with the value function being:
\begin{align}
\begin{split}
\min_{G_k} \max_{D_k} V(D_k, G_k) = &\mathbb{E}_{\mathbf{x}^k\sim p_{C_k}(\mathbf{x}^k)}[\log D_k(\mathbf{x}^k)] + \\
& \mathbb{E}_{\mathbf{z}\sim p_{z}(\mathbf{z})}[\log(1-D_k(G_k(\mathbf{z})))],
\end{split}
\end{align}
where $p_{C_k}(\mathbf{x}^k)$ is the distribution of real load patterns in the $k$-th cluster. In the minimax game, both the generator and discriminator are optimized alternately, where $G_k$ is trained to generate synthetic load patterns that $D_k$ cannot distinguish from the real ones, while $D_k$ is simultaneously trained to improve its ability to distinguish real patterns from generated ones. Formally, we define the loss function for $G_k$ and $D_k$ as:
\begin{align}
\begin{split}
\mathcal{L}_{G_k} = & \mathbb{E}_{\mathbf{z} \sim p(\mathbf{z})}[\log(1 - D_k(G_k(\mathbf{z})))] + \\
& \lambda (|| \mathbb{E}_{\mathbf{z} \sim p(\mathbf{z})}[\mu_{G_k(\mathbf{z})}] - \mathbb{E}_{\mathbf{x}^k \sim p_{C_k}(\mathbf{x}^k)}[\mu_{\mathbf{x}^k}] || + \\
&|| \mathbb{E}_{\mathbf{z} \sim p(\mathbf{z})}[\sigma_{G_k(\mathbf{z})}] - \mathbb{E}_{\mathbf{x}^k \sim p_{C_k}(\mathbf{x}^k)}[\sigma_{\mathbf{x}^k}] ||),
\end{split}
\end{align}

\begin{align}
\begin{split}
\mathcal{L}_{D_k} = &-\mathbb{E}_{\mathbf{x}^k \sim p_{C_k}(\mathbf{x}^k)}[\log D_k(\mathbf{x}^k)] - \\
& \mathbb{E}_{\mathbf{z} \sim p(\mathbf{z})}[\log(1 - D_k(G_k(\mathbf{z})))],
\end{split}
\end{align}
where $\mu_{G(\mathbf{z})}$ and $\sigma_{G(\mathbf{z})}$ represent the mean and variance of the synthetic load pattern generated by the generator $G$; $\mu_{\mathbf{x}}$ and $\sigma_{\mathbf{x}}$ represent the mean and variance of the original load pattern $\mathbf{x}$; $\lambda$ is a factor controlling the importance of the statistical match.

In our work, we set $\lambda$ to a large value (i.e., 100) to heavily emphasize the importance of the statistical match in the generator's optimization process. This choice is due to the application of K-means clustering process, where the residential load patterns within one cluster should have stable statistical properties while still demonstrating temporal variability. Thus, by setting a large value for $\lambda$, we heavily emphasize the generator's ability to match these statistical properties of the real load patterns. This mechanism effectively pushes the generator to not only capture the temporal dynamics but also to replicate the overall statistical characteristics of the real data. In effect, the large $\lambda$ value acts as strong guidance for the generator, ensuring that the synthetic load patterns generated are representative and realistic at both the micro (time-dependent fluctuations) and macro (overall statistical properties) levels. The detailed training hyperparameters are provided in Table \ref{tab:training_parameters} in Section~\ref{sec:result}.

\subsection{Generation of Synthetic Dataset via Ensemble Method}
After training the GANs for each distinct cluster $C_k$, we have effectively obtained a set of generators denoted as $G_k$. Each $G_k$ is proficient in generating synthetic load patterns that effectively capture the distinct characteristics associated with their respective residential load clusters. The final step in this framework involves creating a comprehensive synthetic dataset that encapsulates the wide-ranging diversity present in the original dataset. To achieve this, we employ an ensemble approach for combining the synthetic data generated from each individual recurrent GAN.

Given a desired volume $M$ for the synthetic dataset, we commence by identifying the proportion $\alpha_k$ of the original dataset $\mathcal{D}$ that each cluster $C_k$ constitutes. This can be mathematically expressed as:
\begin{equation}
\alpha_k = \frac{|C_k|}{N}.
\end{equation}

Subsequently, we calculate the number of synthetic load patterns $M_k$ that ought to be generated from each GAN $G_k$. This is achieved by multiplying the desired synthetic dataset volume $M$ by the respective cluster's proportion $\alpha_k$:
\begin{equation}
M_k = \lfloor M \cdot \alpha_k \rceil.
\end{equation}

Following this, we generate $M_k$ synthetic load patterns from each GAN corresponding to each cluster $C_k$, resulting in a synthetic clustered dataset $\hat{C}_k$. This strategic approach ensures that the synthetic dataset $\hat{\mathcal{D}}$ resonates with the original dataset $\mathcal{D}$ in terms of exhibiting a diverse spectrum of load characteristics. Subsequently, we merge the synthetic load patterns generated from all GANs to assemble the final synthetic dataset $\hat{\mathcal{D}}$:
\begin{equation} \label{eq:cluster_gen}
    \hat{C}_{k} = G_k(\mathcal{Z}_k),
\end{equation}
\begin{equation}
\hat{\mathcal{D}} = \bigcup_{k=1}^{K} \hat{C}_{k},
\end{equation}
where $\mathcal{Z}_k$ symbolizes the input noise vector for the $k$-th GAN, with each $\mathcal{Z}_k$ being a matrix with the size of $M_k \times T$.

\section{Benchmark Models and Evaluation Methods}\label{sec:benchmark}

In this section, we present our selection of benchmark models and the corresponding evaluation methodologies employed to assess the performance of synthetic residential load pattern generation. The benchmark models considered in this study include the ERGAN-baseline, Auxiliary Classifier Generative Adversarial Network (ACGAN), Wasserstein Generative Adversarial Network (WGAN), and Continuous RNN-GAN (C-RNN-GAN), each providing unique attributes and benefits relevant to our problem setting. We provide the rationale behind these selections and highlight their differences compared to ERGAN. Subsequently, we present the evaluation methods deployed to assess the quality and diversity of the synthetic load patterns, facilitating a comprehensive and interpretable comparison between ERGAN relative and above-mentioned benchmark models.

\subsection{Benchmark Models}

In this study, we perform a comparative evaluation of our proposed ERGAN framework alongside state-of-the-art models, providing a robust assessment of ERGAN's performance in generating realistic synthetic residential load patterns. The selected benchmark models for comparison covers a range of techniques that have demonstrated considerable promise in the task of generative modeling, particularly time-series data generation. Each of these models brings unique capabilities, providing a robust basis for comparison with our proposed ERGAN framework. Details are discussed as follows.
\begin{itemize}
    \item \textbf{ERGAN-Baseline} serves as a benchmark model in our study, representing the ERGAN framework without the K-means clustering and ensemble methods. By comparing its performance against the complete ERGAN framework and other benchmark models, we can evaluate the added value and effectiveness of techniques developed in this paper, such as clustering and ensemble methods,  in generating diverse and realistic synthetic residential load patterns.
    \item \textbf{Auxiliary Classifier GAN (ACGAN)} \cite{odena2017conditional} extends the traditional GAN structure by conditioning both the generator and the discriminator on class labels, thereby enabling the generation of specific classes of data. It has been previously utilized in a similar context for residential load generation \cite{gu2019gan}, showing its suitability as a benchmark for our study. Their conditioning of ACGAN is similar to our use of cluster information in generating load patterns, although the methods differ substantially. ACGAN is a form of conditional GAN where the same model is used for all categories, conditioned on the category label. In contrast, our ERGAN framework employs an ensemble of GANs, each tailored to a specific cluster, and thus being capable of capturing distinct characteristics across clusters. However, despite its merits, ACGAN uses convolutional neural network (CNN) which does not incorporate the advantages of Bi-LSTM in handling time-series data and does not focus on the preservation of statistical properties, as our framework does.
    \item \textbf{Wasserstein GAN (WGAN)} \cite{arjovsky2017wasserstein} provides an alternative approach to the traditional GAN loss function to address the issue of training instability, which is known as a common challenge in training GAN models. Its notable characteristic of improved training stability and convergence offers a valuable benchmark for comparison with our ERGAN model, which also emphasizes stable and efficient training. Nevertheless, WGAN also uses CNN which might not inherently accommodate the global temporal dependencies present in our load pattern data as well as our approach does.
    \item \textbf{C-RNN-GAN} \cite{mogren2016c} is a robust model for time-series generation, harnessing the power of recurrent neural networks (RNN) and adversarial training from GANs. It uses LSTM units, ideal for learning and remembering long-term dependencies in the data, which are vital in the context of residential load patterns. Despite these strengths, the C-RNN-GAN does not explicitly cater to the diversity observed in residential load patterns across different households, a feature our ensemble of Bi-LSTM GANs addresses. Furthermore, C-RNN-GAN, although using Bi-LSTM units in the discriminator, does not incorporate a bidirectional structure in its generator which lacks consideration of the time-series data forward.
\end{itemize}
By comparing our ERGAN framework with these diverse benchmark models, we aim to thoroughly evaluate and validate the performance and effectiveness of our approach in generating synthetic residential load patterns.

\subsection{Evaluation Methods}
The quality of the synthetic residential load patterns generated by our ERGAN framework and the selected benchmark models are evaluated via three distinct but complementary methods, with details described as follows.

\begin{itemize}
    \item {Visual Examination of Real and Synthetic Load Patterns and Their Autocorrelation:} We randomly select multiple real load pattern samples and compare them to the samples generated by different models. Each generated sample is selected to match the real samples based on the minimum Euclidean distance. Additionally, we employ auto-correlation techniques to examine correlations between observations at different points in the time series, enriching our analysis of similarities and differences. 
    \item \textbf{Comparative Histograms of Original and Synthetic Residential Load Patterns:} This metric employed histogram-based comparison to visualize the distributional attributes of original and synthetic data. In this approach, all load patterns produced by a generative model are aggregated together, and its histogram is plotted alongside that of the original data. By investigating to what extent the histogram of generated synthetic load patterns aligns with that of the original data, we can assess the model's ability to accurately replicate the data's global statistical properties.
    \item \textbf{Hourly Comparative Boxplots for Original and Synthetic Load Patterns:} Our evaluation methodology further incorporates a time-centric examination of the generated load patterns. Owing to the data's hourly nature, we create box plots for each time slot, showcasing the distribution of generated and original data per time step. This approach can provide insights into the model's capacity to mimic the temporal fluctuations of residential load patterns.
    \item \textbf{Comparative Visualization of T-SNE Dimension-Reduced Load Patterns} Another component of our evaluation method is the use of t-SNE visualizations. This high-dimensional data visualization technique enables us to compare the manifold structures of original and synthetic load patterns. The closeness of synthetic data points to those of the original data in this reduced-dimension space indicates the model's proficiency in preserving the manifold structure of high-dimensional load patterns.
    \item \textbf{Quantitative Evaluation using Statistical Distances:} The final component of our evaluation method is to assess the statistical similarity between the original and synthetic datasets is by comparing key statistical properties: the mean, variance, 25th percentile (Q1), and 75th percentile (Q3) profiles. The L1 distance (sum of absolute differences) is calculated for each property's profile between the original and synthetic datasets. The model demonstrating the lowest L1 distances across these properties is deemed to have generated synthetic data that most closely aligns with the statistical characteristics of the real data.
\end{itemize}

These evaluation methods offer comprehensive insights into the generative models' performance, assessing their ability to accurately learn and capture the original data's statistical, temporal, and structural properties.

\section{Result and Analysis}\label{sec:result}
This section firstly provides an in-depth description of the residential load pattern data used in this paper. Following this, we present a comprehensive performance evaluation using the metrics described in Section~\ref{sec:benchmark}, including the real and synthetic sample sets visualized in pattern and autocorrelation 
depicted in Figure \ref{fig:pattern&autocorrelation}, comparative histograms depicted in Figure \ref{fig:histogram}, hourly comparative boxplots depicted in Figure \ref{fig:boxplot}, and comparative visualization of T-SNE dimension-reduced load patterns depicted in Figure \ref{fig:scatter_plot}. Through these robust assessments, we aim to demonstrate the effectiveness of our ERGAN framework in generating synthetic residential load patterns that accurately capture the statistical, temporal, and structural properties of real-world data.

\subsection{Data Description and Model Setup}

\begin{table}[t]
\centering
\caption{ERGAN Model Architecture}\label{tab:model_architecture}
\begin{tabular}{|c|c|}
\hline
\textbf{Component} & \textbf{Description} \\ 
\hline
\multicolumn{2}{|c|}{\textbf{Generator}} \\ 
\hline
LSTM Layer & 1 input, 16 hidden units, 5 layers, bidirectional \\ 
\hline
Fully Connected Layer & 32 inputs, 1 output \\ 
\hline
\multicolumn{2}{|c|}{\textbf{Discriminator}} \\ 
\hline
LSTM Layer & 1 input, 16 hidden units, 5 layers, bidirectional \\ 
\hline
Fully Connected Layer & Linear layer (64 inputs, 1 output) \\ 
\hline
\end{tabular}
\end{table}

\begin{table}[t]
\centering
\caption{Hyperparameters for Model Training}\label{tab:training_parameters}
\begin{tabular}{cc}
\hline
\hspace{1cm} \textbf{Parameters} & \hspace{1cm} \textbf{Details} \hspace{1cm} \\
\hline
\hspace{1cm} Epochs & \hspace{1cm} 10000 \hspace{1cm} \\
\hspace{1cm} Batch size & \hspace{1cm} 1024 \hspace{1cm} \\
\hspace{1cm} Optimizer & \hspace{1cm}Adam \hspace{1cm} \\
\hspace{1cm} Generator Learning rate & \hspace{1cm} 0.0001 \hspace{1cm} \\
\hspace{1cm} Discriminator Learning rate & \hspace{1cm} 0.0001 \hspace{1cm} \\
\hline
\end{tabular}
\end{table}

In this study, we use the Pecan Street dataset~\cite{WinNT}, which provides hourly residential energy consumption data from 417 households for the entire year from January 1, 2017 to December 31, 2017. This dataset contains hourly smart meter readings, and we specifically use total household energy consumption from all electrical sources within each household.
    
While our model does not explicitly separate the data by day types or seasons, the inclusion of data spanning all days of the week and all seasons ensures that our synthetic load patterns reflect the inherent variability associated with different times of the year. To ensure the integrity of our analysis, each 24-hour load profile is segmented from the smart meter data, and all sequences containing missing values are prudently removed. Subsequently, to concentrate on load patterns rather than the magnitude of electricity consumption, each sequence is normalized using linear scaling. This normalization process ensures that each sequence falls within the range $[0,1]$, allowing for consistent analysis and comparison across various households. The well-curated dataset is then partitioned into training, validation, and testing subsets, with 70\% of the data dedicated to training, and the remaining 30\% for validation.

For our proposed ERGAN model, we present its model architecture in Table~\ref{tab:model_architecture} and training hyperparameter setting in Table~\ref{tab:training_parameters}, respectively. 

\subsection{Number of Cluster Selection and its Impact}
\begin{figure}[t]
\centerline{\includegraphics[width=\columnwidth]{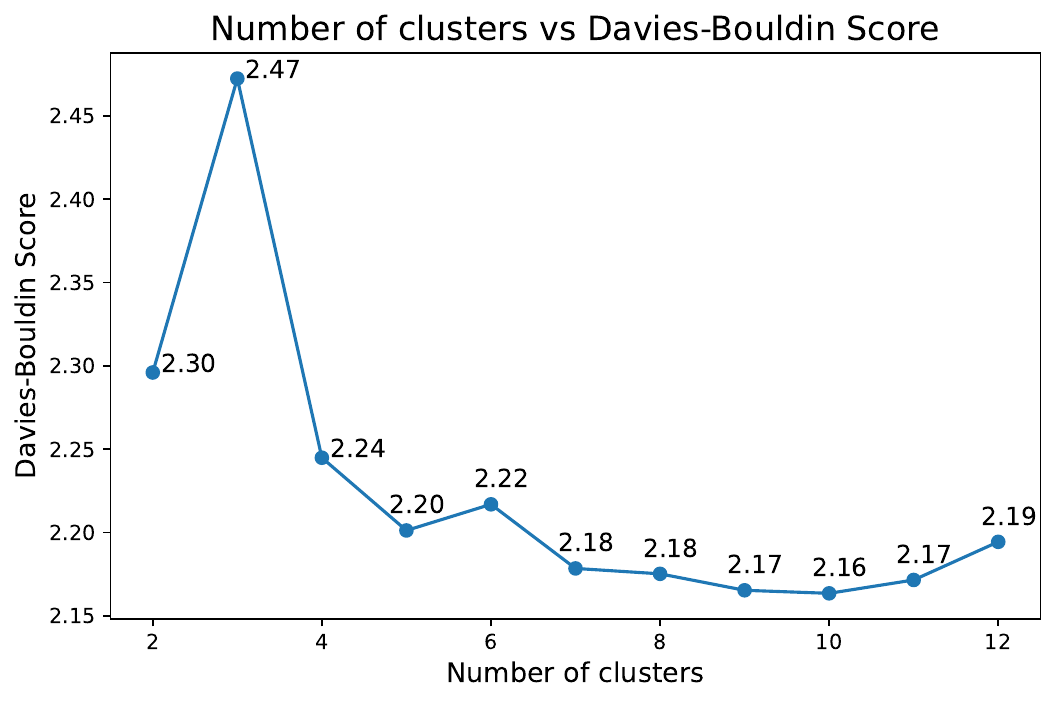}}
\caption{Selecting the best number of clusters based on the Davies-Bouldin score.}
\label{fig:kmeans}
\end{figure} 
In the ERGAN framework, the choice of the number of clusters $K$ is a key factor that influences both the quality of synthetic data and the computational efficiency of the model. The Davies-Bouldin (DB) index is employed to assess clustering quality by evaluating the separation and cohesion of clusters, where a lower DB index generally suggests a better configuration for capturing residential load patterns. However, it is important to note that there is no definitively optimal $K$; the choice of $K$ is inherently dependent on the method used and the specific characteristics of the dataset. While the DB index provides useful guidance, the primary goal is to generate synthetic data that reflects the diversity and complexity of real-world residential loads while managing computational overhead.
    
Choosing $K$ involves balancing the benefits of a more detailed data representation against the challenges of increased computational demands. For example, a larger $K$ can offer a more detailed representation of the data distribution, potentially enhancing the diversity and realism of the synthetic data. However, such a larger $K$ also requires training more generative models, leading to increased computational overhead and a greater risk of overfitting, where models may capture noises rather than meaningful patterns. Additionally, too many clusters can fragment coherent data structures, impacting the interpretability and quality of the generated data. Conversely, selecting a smaller $K$ may reduce computational costs but risks underfitting, where the model oversimplifies the data and fails to capture its variability. This can result in synthetic data that lacks the necessary diversity to accurately represent residential energy consumption patterns. To balance these trade-offs, we calculated the DB index for cluster sizes from 2 to 12 as shown in Figure 2. The results suggest that $K=10$ provides a practical balance, effectively capturing diverse load patterns without excessive computational costs or sacrificing data quality. This choice aligns with our study's objectives and offers a reasonable balance between complexity, efficiency, and quality. The effectiveness of this clustering strategy is evident in the ERGAN model's performance, which shows better results than benchmark models in capturing the statistical and temporal characteristics of residential load data. The detailed performance comparisons and results of ERGAN with $K=10$ are discussed in the subsequent sections.

\subsection{Results Analysis and Insights}

\begin{figure*}[htbp]
\centering
\includegraphics[width=\textwidth]{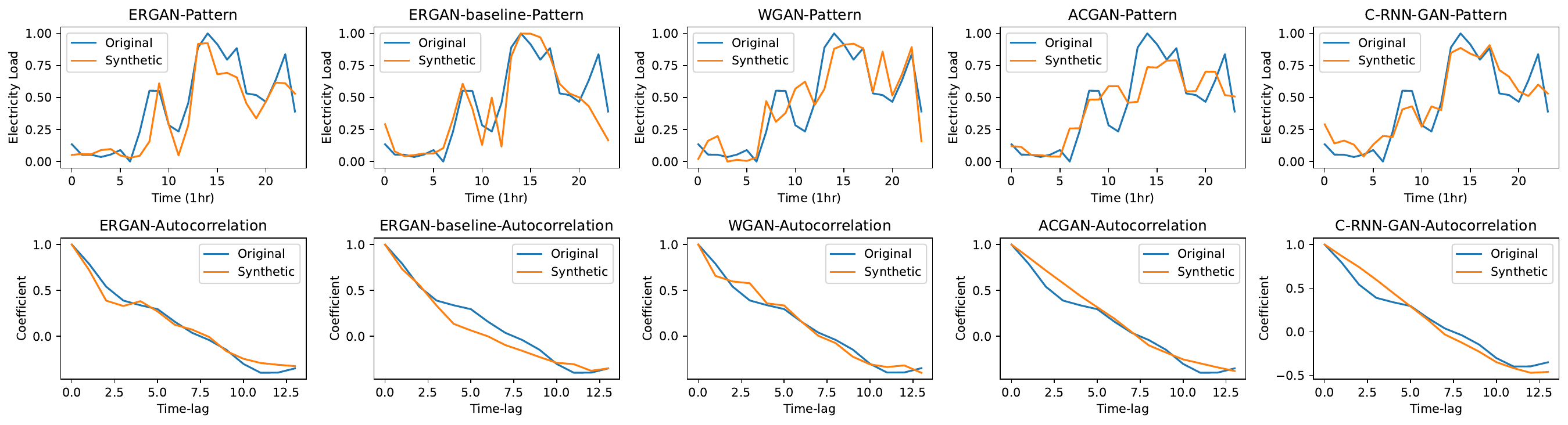}
\includegraphics[width=\textwidth]{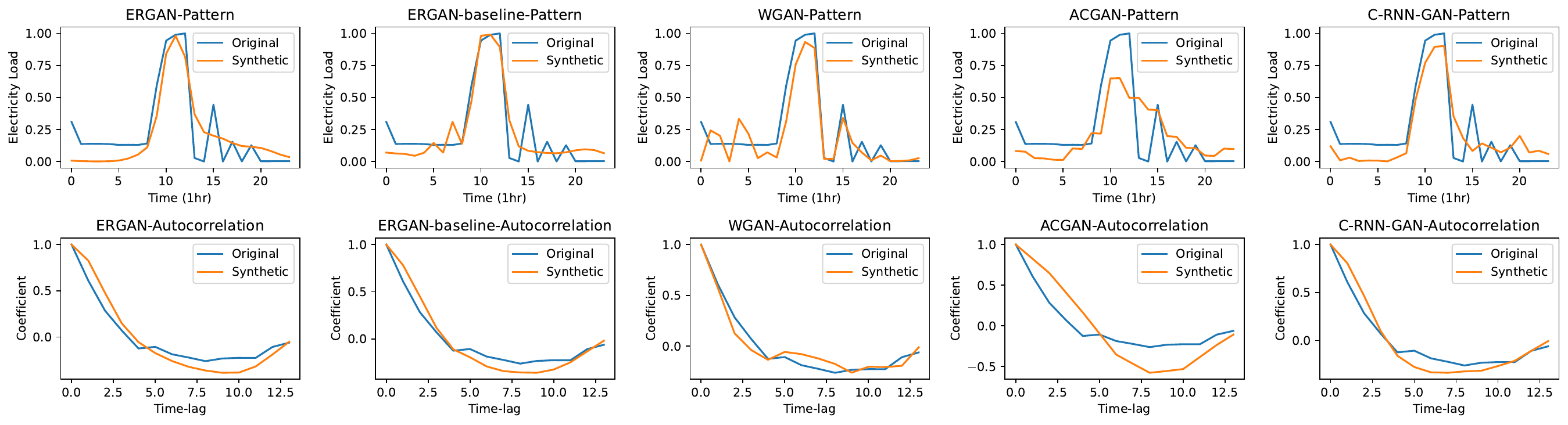}
\includegraphics[width=\textwidth]{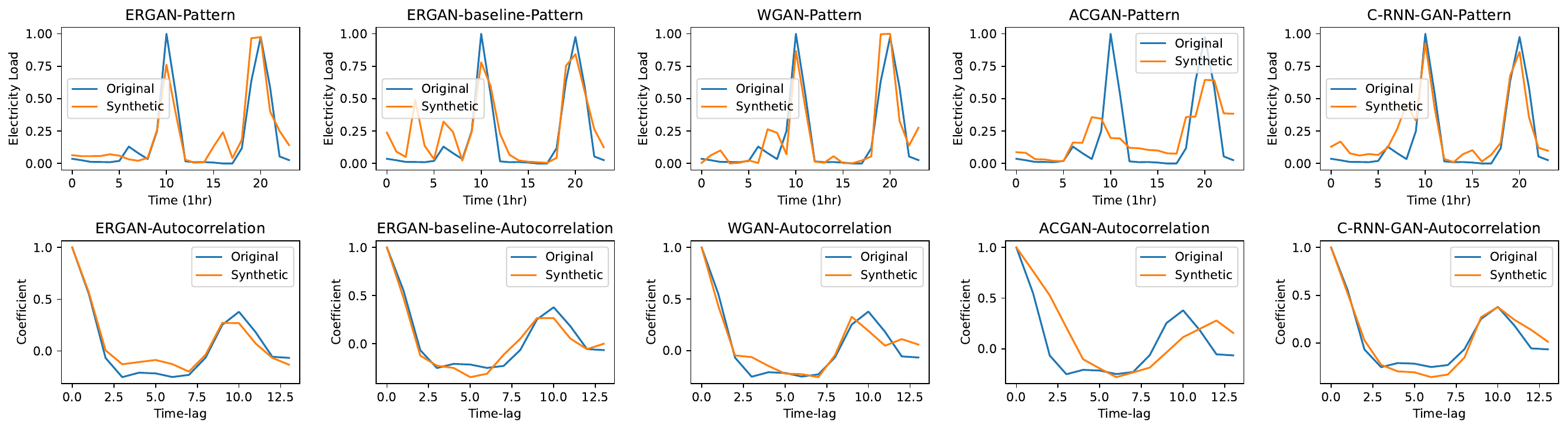}
\includegraphics[width=\textwidth]{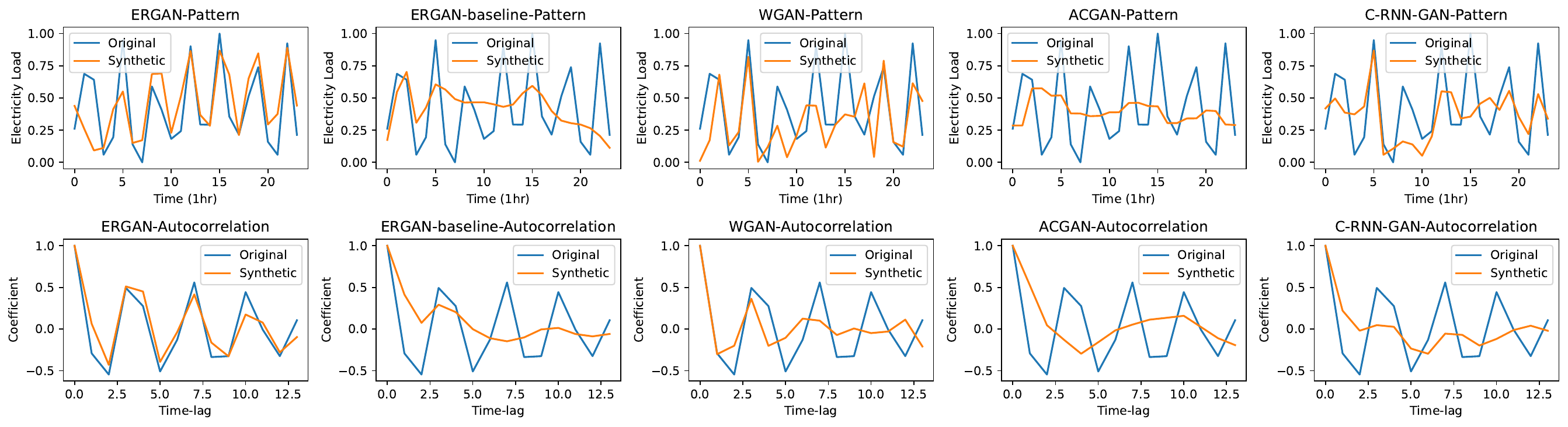}
\captionsetup{justification=centering}
\caption{Pattern and correlated autocorrelation comparison of original and synthetic residential load patterns via different generation methods.}
\label{fig:pattern&autocorrelation}
\end{figure*}

\begin{figure*}[htbp]
\centering
\includegraphics[width=\textwidth]{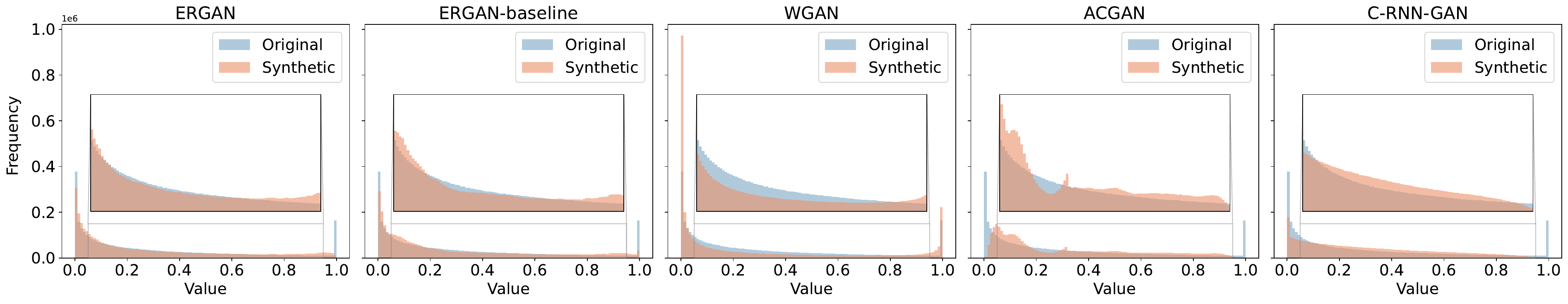}
\captionsetup{justification=centering}
\caption{Comparative histograms of original and synthetic residential load patterns via different generation methods.}
\label{fig:histogram}
\end{figure*} 

\begin{figure*}[htbp]
\centering
\includegraphics[width=\textwidth]{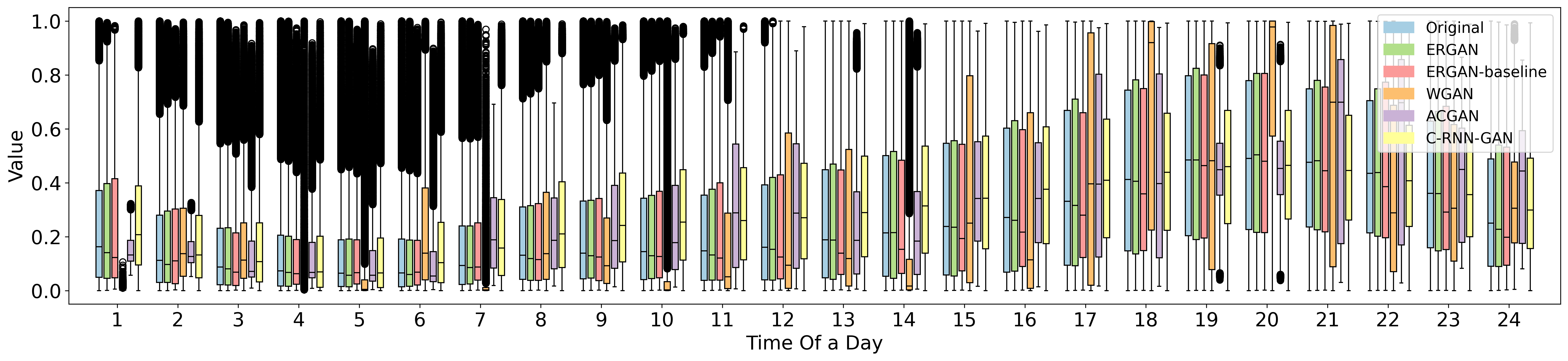}
\captionsetup{justification=centering}
\caption{Hourly comparative boxplots for original and synthetic load patterns generated by ERGAN and benchmark methods.}
\label{fig:boxplot}
\end{figure*} 

\begin{figure*}[htbp]
\centering
\includegraphics[width=\textwidth]{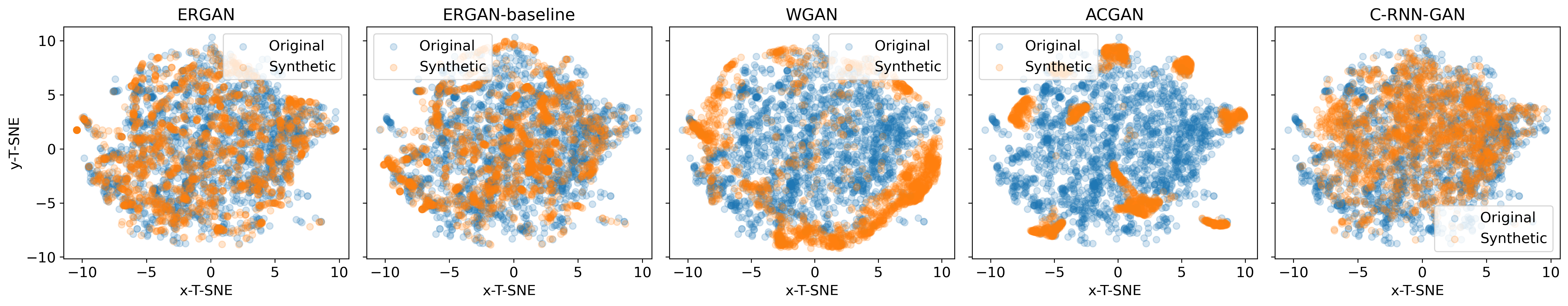}
\captionsetup{justification=centering}
\caption{Comparative visualization of T-SNE dimension-reduced load patterns using original against synthetic data across multiple generation methods.}
\label{fig:scatter_plot}
\end{figure*} 

\begin{table}[ht]
\centering
\caption{L1-Based Distance Analysis}
\label{tab:dist_measurement}
\begin{tabular}{lcccc}
\hline
\textbf{Model} & \textbf{Mean L1} & \textbf{Variance L1} & \textbf{Q1 L1}  & \textbf{Q3 L1}  \\ \hline
\textbf{ERGAN} & \textbf{0.0333} & \textbf{0.0138} & \textbf{0.1186} & \textbf{0.4885} \\
ERGAN-baseline & 0.1775 & 0.0297 & 0.2258 & 0.4909 \\
WGAN & 2.0767 & 0.8038 & 1.5132 & 3.7942 \\
ACGAN & 1.1032 & 0.6305 & 1.3049 & 2.3666 \\
C-RNN-GAN & 0.8328 & 0.5972 & 1.2641 & 1.4356 \\ \hline
\end{tabular}
\end{table}

\subsubsection{Outstanding Performance of ERGAN over Benchmark Models in Ensuring Diversity and Similarity in Synthetic Load Patterns}

The effectiveness of ERGAN in creating synthetic residential load patterns is affirmed through an comprehensive comparison with the benchmark models. Figure \ref{fig:pattern&autocorrelation} reveals that all models, including ERGAN, are capable of identifying certain load patterns inherent to the original data. Despite this common capability, ERGAN consistently outperforms the benchmark models in terms of capturing the statistical properties of the real data, as evidenced by its significantly lower L1 distances across all four metrics (mean, variance, Q1, and Q3) shown in Table \ref{tab:dist_measurement}.

The comparative histograms shown in Figure \ref{fig:histogram} reveal the diverse range of patterns generated by our ERGAN method. A well-aligned spread is observed, matching closely with the original data's global statistical attributes and thus indicating a comprehensive replication of the overall load consumption behavior. In contrast, the benchmark models display slight deviations from the original data distribution, suggesting potential challenges in accurately capturing the intricate statistical characteristics. 

The hourly comparative boxplots shown in Figure \ref{fig:boxplot} further reinforce these observations. Our ERGAN method exhibits the ability to replicate temporal fluctuations across the day convincingly, evident by closely matching the spread of the original data for each hourly time slot. Meanwhile, other benchmark models seem to struggle with capturing data distribution at certain time steps, reflecting reduced quality and diversity of the generated load patterns. 

Furthermore, the t-SNE visualization of the original and synthetic load patterns shown in Figure \ref{fig:scatter_plot} also demonstrates the effectiveness of our methodology. The generated load patterns, when projected in a lower-dimensional space, form data points that closely align with those formed by the original data, implying a strong preservation of the high-dimensional manifold structure. On the contrary, the benchmark models show a divergence in this aspect, with their synthetic data points presenting a higher level of discrepancy, or in some cases overly concentrated.

\subsubsection{Enhanced Performance of Recurrent GAN Models over CNN GAN Models}

The evaluation of generative models reveals the superiority of RNN models, specifically the ERGAN, ERGAN-baseline, and C-RNN-GAN models, over the CNN based models, such as WGAN and ACGAN. This finding is supported by the observed characteristics in the evaluation plots. Specifically, the comparative histograms shown in Figure \ref{fig:histogram} demonstrate that the synthetic load patterns generated by the recurrent GAN models in ERGAN closely align with the distributional attributes of the original data, indicating their proficiency in replicating the global statistical properties. The quantitative results, presented in Table \ref{tab:dist_measurement}, further solidify this observation. The RNN-based models, including ERGAN, ERGAN-baseline, and C-RNN-GAN, exhibit considerably lower L1 distances compared to WGAN and ACGAN, indicating their superior ability to replicate the statistical characteristics of the real data. Moreover, the hourly boxplots shown in Figure \ref{fig:boxplot} showcase the ability of the recurrent models to capture the temporal fluctuations present in residential load patterns. The close resemblance of the boxplots for each time slot between the synthetic and original data further reinforces the recurrent models' capability in mimicking the temporal dynamics. Lastly, the t-SNE visualizations shown in Figure \ref{fig:scatter_plot} exhibit the close proximity of the synthetic data points to the original data points in the reduced-dimension space, indicating the recurrent models' effectiveness in preserving the manifold structure of the high-dimensional load patterns.

\subsubsection{Utilizing Bi-LSTM Throughout the Architecture and Leveraging Statistical Properties in the Loss Function for Enhanced Residential Load Pattern Generation}

The integration of LSTM units as employed by C-RNN-GAN and ERGAN-baseline is a promising strategy due to the inherent capability of LSTM in handling long-term dependencies in residential load patterns. While both models leverage LSTM units, their fundamental differences lie in the directionality of LSTM implementation and the consideration of statistical properties in the loss function, facilitating an insightful comparative study. Specifically, the ERGAN-baseline model utilizes Bi-LSTM units throughout its architecture and incorporates statistical properties in the loss function, unlike C-RNN-GAN which uses LSTM in a unidirectional manner in its generator and solely relies on adversarial loss.

Insights drawn from our comparative study reveal that ERGAN-baseline consistently outperforms C-RNN-GAN across all evaluation metrics, underscoring the significance of bidirectional LSTM and statistical property-focused loss function. The quantitative results shown in Table \ref{tab:dist_measurement} provide evidence of this superior performance. The lower L1 distances achieved by ERGAn and ERGAN-baseline across all metrics underscore the benefits of employing Bi-LSTM throughout the architecture and incorporating statistical properties in the loss function. The comparative histograms shown in Figure \ref{fig:histogram} demonstrate a closer alignment of synthetic load patterns generated by ERGAN-baseline to the original data distribution, indicating its superior capacity in capturing intricate statistical characteristics. Similarly, the hourly comparative boxplots depicted in Figure \ref{fig:boxplot} shows ERGAN-baseline's ability to faithfully replicate the temporal fluctuations in load patterns, with its synthetic data more closely matching the spread of the original data across different hourly time slots.

The additional ensemble method deployed in the ERGAN framework is not included in the ERGAN-baseline model, making the comparison against C-RNN-GAN fairer. Despite the absence of this enhancement, the performance of the ERGAN-baseline still surpasses the C-RNN-GAN, demonstrating the effectiveness of employing Bi-LSTM throughout the architecture and incorporating statistical properties in the loss function. The resulting synthetic residential load patterns exhibit improved diversity, realism, and temporal accuracy, establishing the merits of such an approach in the domain of residential load pattern generation.

\subsubsection{Effectiveness of the Ensemble Approach in Enhancing the Diversity and Quality of Synthetic Load Patterns}

The ensemble approach incorporated in our ERGAN framework can enhance the diversity and quality of the generated synthetic residential load patterns. By comparing ERGAN with its ERGAN-baseline counterpart, where the ensemble approach is absent, we can delineate the substantial improvements brought about by the ensemble strategy. Comparative histograms shown in Figure \ref{fig:histogram} and hourly boxplots shown in Figure \ref{fig:boxplot} indicate ERGAN's superior ability to capture the intricate global statistical characteristics and the temporal fluctuations of the load patterns more accurately than its baseline model. These improvements can also be found in the t-SNE scatter plot visualization shown in Figure \ref{fig:scatter_plot}, where ERGAN exhibits a more accurate preservation of the manifold structure of the high-dimensional load patterns. This improvement is not only evident in the qualitative evaluations but is also supported by the quantitative results. As shown in Table \ref{tab:dist_measurement}, ERGAN achieves even lower L1 distances compared to its baseline counterpart (ERGAN-baseline), demonstrating the quantitative benefits of the ensemble strategy in capturing the statistical properties of the real data. Through a strategy that encourages diversity among individual generators and effectively amalgamates their strengths, the Ensemble approach significantly enhances the realism, diversity, and temporal accuracy of the synthetic data, validating the effectiveness of this technique in the ERGAN framework.

\section{Conclusion}
This study addressed the vital task of synthetic residential load pattern generation by introducing an innovative framework-the Ensemble Recurrent Generative Adversarial Network Framework called ERGAN. The proposed framework employs an ensemble of recurrent GANs and a unique loss function incorporating statistical properties. Through comparative analysis with various state-of-the-art benchmarks, ERGAN has proven its superior performance in generating diverse and high-quality synthetic load patterns, closely emulating real-world scenarios. The study can contribute to various applications in power system operation and energy management, where load data is needed. Yet, opportunities for further exploration remain, including the potential to optimize the ERGAN model by incorporating advanced feature conditions for time-dependent residential load demand data generation. Thus, this research represents a step forward in efficient and realistic residential load data generation.

\bibliographystyle{IEEEtran}
\bibliography{references}

\begin{IEEEbiography}
[{\includegraphics[width=1in,height=1.25in,clip,keepaspectratio]{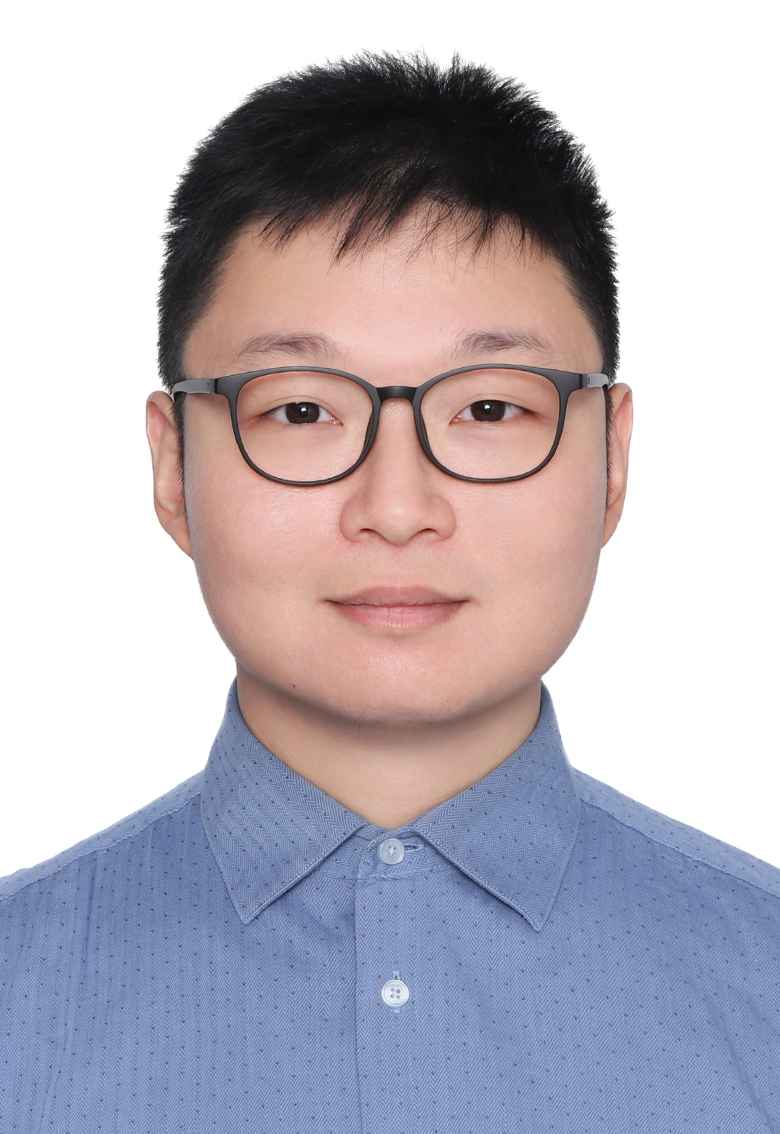}}]{Xinyu Liang} received his Bachelor of Computer Science and Master of Artificial Intelligence degrees from Monash University, Melbourne, VIC, Australia, in 2018 and 2021, respectively. He is currently pursuing his Ph.D. degree in Information Technology in the Department of Data Science and AI, Faculty of Information Technology, Monash University, Melbourne, VIC, Australia. His research interests include energy informatics and human-in-the-loop AI for smart energy systems.
\end{IEEEbiography}

\begin{IEEEbiography}
[{\includegraphics[width=1in,height=1.25in,clip,keepaspectratio]{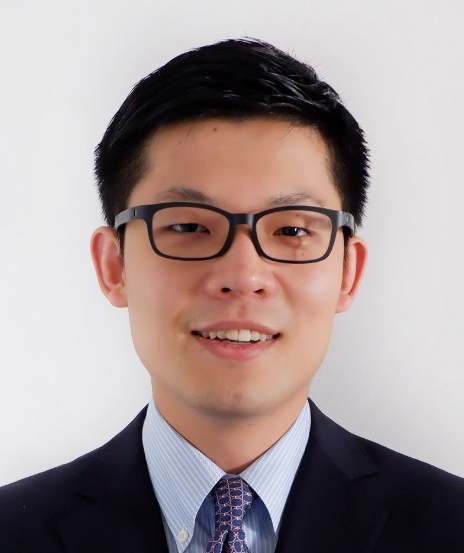}}]{Ziheng Wang} received his Master's degree in Applied Mathematics from the State University of New York at Stony Brook, NY, USA, in 2018 and M.Sc. in Information Technology from Monash University, Melbourne, VIC, Australia in 2021. He is currently working as a data scientist, specializing in marketing analytics within the banking sector. His research interests include data driven models with business applications.
\end{IEEEbiography}

\begin{IEEEbiography}
[{\includegraphics[width=1in,height=1.25in,clip,keepaspectratio]{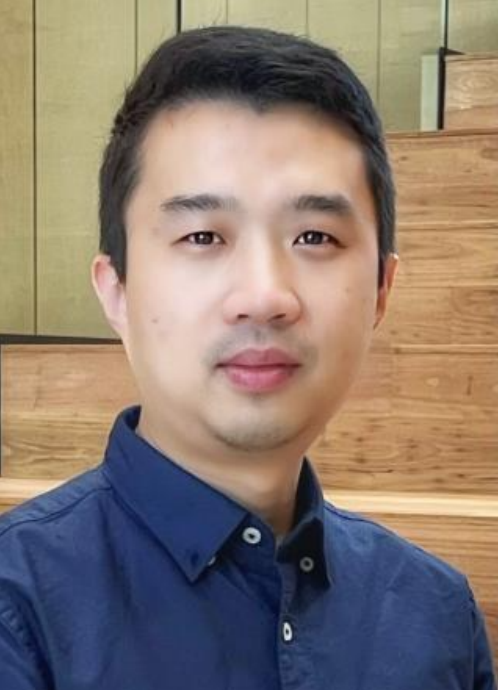}}]{Hao Wang} (M'16) received his Ph.D. in Information Engineering from The Chinese University of Hong Kong, Hong Kong, in 2016. He was a Postdoctoral Research Fellow at Stanford University, Stanford, CA, USA, and a Washington Research Foundation Innovation Fellow at the University of Washington, Seattle, WA, USA. He is currently a Senior Lecturer and ARC DECRA Fellow in the Department of Data Science and AI, Faculty of IT, Monash University, Melbourne, VIC, Australia. His research interests include optimization, machine learning, and data analytics for power and energy systems.
\end{IEEEbiography}

\end{document}